\documentclass[journal]{IEEEtran}

\usepackage{soul,framed} 
\usepackage[table]{xcolor}
\colorlet{shadecolor}{yellow}
\usepackage{graphicx}
\graphicspath{{Fig/}}
\DeclareGraphicsExtensions{.pdf,.png,.jpg,.jpeg}

\usepackage{marvosym}
\usepackage[cmex10]{amsmath}
\usepackage{array}
\usepackage{mdwmath}
\usepackage{mdwtab}
\usepackage{eqparbox}
\usepackage{url}
\usepackage{amsfonts}
\usepackage[ruled,vlined]{algorithm2e}
\usepackage{multirow}
\usepackage{footnote}
\usepackage{bm}
\usepackage{makecell}
\usepackage{pdflscape}
\usepackage{afterpage}
\usepackage{lscape}
\usepackage{booktabs}
\usepackage{placeins}
\usepackage{float}
\usepackage{tikz}

\usepackage{pgfplots}
\pgfplotsset{compat=1.12}
  \pgfplotsset{
    layers/my layer set/.define layer set={
      background,
      main,
      foreground
    }{ },
    set layers=my layer set,
  }

\usepackage[square,sort,comma,numbers]{natbib}

\usepackage[pagebackref=false,breaklinks=false,linkcolor=black,anchorcolor=black,citecolor=blue,urlcolor=blue,colorlinks,bookmarks=true]{hyperref}

\usepackage{pifont}

\definecolor{mygray}{gray}{.55}
\definecolor{mypink}{RGB}{244,0,122}
\definecolor{mygreen}{RGB}{0,104,0}
\definecolor{myyellow}{RGB}{255,102,0}
\definecolor{DarkBlue}{rgb}{0,0,1}
\usepackage{amssymb}
\usepackage{colortbl}

\newcommand{\TableFont}{\fontsize{8pt}{9.4pt}\selectfont}
\begin{document}
	
\title{MI-DETR: A Strong Baseline for Moving Infrared Small Target Detection with Motion Integration}

\author{
    Nian~Liu$^{*}$,
    Jin~Gao$^{*}$,
    Zhen~Liang,
    Shubo~Lin,
    Sikui~Zhang,
    Fudong~Ge,
    Liang~Li,
    and Weiming~Hu,~\IEEEmembership{Senior Member,~IEEE}%
    \thanks{$^{*}$Nian Liu and Jin Gao contributed equally to this work. Corresponding author: Jin Gao (jin.gao@nlpr.ia.ac.cn).}%
    \thanks{N. Liu is with the School of Advanced Interdisciplinary Sciences (SAIS), University of Chinese Academy of Sciences (UCAS), Beijing 101408, China, and also with the State Key Laboratory of Multimodal Artificial Intelligence Systems, Institute of Automation, Chinese Academy of Sciences, Beijing 100190, China.}%
    \thanks{J. Gao, S. Lin, S. Zhang, F. Ge, and W. Hu are with the State Key Laboratory of Multimodal Artificial Intelligence Systems, Institute of Automation, Chinese Academy of Sciences, Beijing 100190, China, and also with the School of Artificial Intelligence, University of Chinese Academy of Sciences, Beijing 101408, China. W. Hu is also with the School of Information Science and Technology, ShanghaiTech University, Shanghai 201210, China.}%
    \thanks{Z. Liang is with the Institute of Optics and Electronics, Chinese Academy of Sciences, Chengdu 610209, China, and also with the University of Chinese Academy of Sciences (UCAS), Beijing 100049, China.}%
    \thanks{L. Li is with the Beijing Institute of Basic Medical Sciences, Beijing 100850, China.}%
}

\maketitle
\begin{abstract}
Detecting moving infrared small targets is challenging because tiny, low-contrast targets occupy few pixels and are easily obscured by dynamic backgrounds. Existing multi-frame methods aggregate temporal information across frames to capture motion. However, dynamic background changes can generate similar motion cues, making it difficult to distinguish between target motion and background interference. Furthermore, even when motion cues are extracted, combining them with current-frame appearance features remains difficult. To address these issues, we propose Motion Integration DETR (MI-DETR), a three-stage framework that explicitly models motion and fuses it with appearance features. First, to suppress background clutter while preserving target-related motion cues, Recurrent Interpretable Motion Cue Aggregation (RIMCA) maintains a recurrent temporal state that accumulates motion across consecutive frames, producing a causal and spatially aligned motion representation. Second, to integrate spatial and temporal information, Pathway Mutual Interaction (PMI) preserves separate appearance and motion pathways while enabling bidirectional feature exchange between them. Finally, an RT-DETR-based detector uses these refined features for end-to-end target localization. Experiments on DAUB-R, ITSDT-15K, and IRDST-H show that explicit motion modeling and pathway interaction effectively improve moving infrared small target detection.
Our code is available at \url{https://github.com/nliu-25/MI-DETR}.

\end{abstract}

\section{Introduction}\label{sec:introduction}

\IEEEPARstart{M}{oving} infrared small target detection aims to localize weak moving targets in infrared video. It is important for long-range aerial monitoring, surveillance, early warning, and related remote-sensing applications~\cite{zhao2022survey,kou2023survey,ying2025rgbttiny}. This task is challenging because distant targets often occupy only a few pixels and have low contrast with the surrounding background. Their shape and texture cues are also limited~\cite{DNAnetli2022dense,zhang2022isnet,chen2023augtarget}. Complex and dynamic backgrounds can further obscure these weak targets. As a result, the appearance information available in a single infrared image is often limited. For moving targets, consecutive frames provide additional temporal information that can be used to capture motion.

\begin{figure}[!t]
\centering
\vspace{-20pt}
\includegraphics[trim=0 20mm 0 3mm, clip,width=0.90\linewidth]{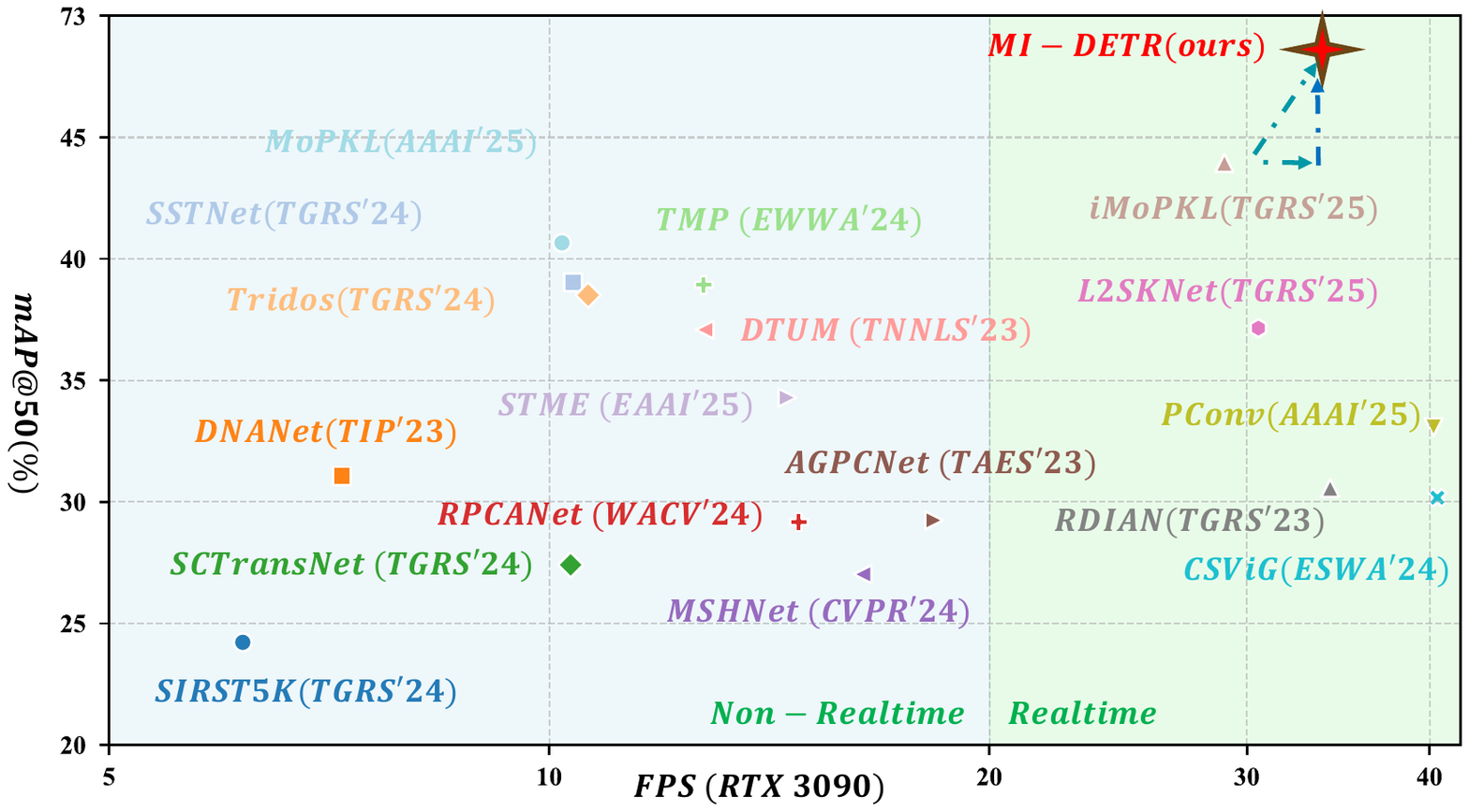}
\caption{MI-DETR combines 70.30\% mAP$_{50}$ with 34.60 FPS on IRDST-H under the unified RTX~3090 protocol~\cite{chen2025motion}.}
\label{fig:accuracy_comparison}
\end{figure}

Existing ISTD methods can be broadly divided into model-driven and data-driven approaches~\cite{zhao2022survey,kou2023survey}. Model-driven methods use hand-crafted features and predefined priors to separate small targets from the background. Representative methods include TopHat~\cite{tom1993morphology}, LCM~\cite{chen2013lcm}, IPI~\cite{gao2013infrared}, and their variants. However, their performance often depends on prior assumptions about the target and background. Such assumptions may not hold in complex infrared scenes, which limits the adaptability of model-driven methods~\cite{zhao2022survey,kou2023survey,DNAnetli2022dense}.

Data-driven methods learn target features directly from data and reduce the dependence on hand-crafted priors. According to the number of input frames, these methods can be further divided into single-frame and multi-frame methods~\cite{kou2023survey}. Single-frame methods, such as ACM~\cite{dai2021asymmetric}, DNA-Net~\cite{DNAnetli2022dense}, ISNet~\cite{zhang2022isnet}, and UIU-Net~\cite{wu2022uiu}, detect small targets from a single infrared image. They mainly use spatial information and avoid the additional computation required for temporal modeling. However, using only one frame also means that temporal information across frames is not used. This limits the available information for moving targets, especially when their appearance cues are weak.

Multi-frame methods address this limitation by using consecutive infrared frames~\cite{du2021spatial,chen2024sstnet,huang2024lmaformer,zhu2024tmp,zhang2025mocid,fang2026tdcnet}. They aggregate temporal information across frames through recurrent modeling, spatio-temporal feature learning, attention, or motion-aware modeling. The temporal information provides motion cues that complement the appearance information of a single frame. However, these motion cues are not always produced by the target. Dynamic background changes can also produce similar motion cues, making target motion difficult to distinguish from background interference. This problem has been studied in recent multi-frame methods. LMAFormer considers interference from moving backgrounds~\cite{huang2024lmaformer}, MOCID models motion and displacement differences between targets and backgrounds~\cite{zhang2025mocid}, and TDCNet suppresses pseudo-motion clutter in complex backgrounds~\cite{fang2026tdcnet}. These studies show that extracting reliable target-related motion cues remains difficult in complex and dynamic scenes.

Even when motion cues are extracted, they still need to be combined with the appearance features of the current frame. Motion cues describe changes across consecutive frames, while appearance features describe the spatial content of the current infrared image. Multi-frame methods have therefore introduced different fusion and interaction mechanisms to combine these two types of information~\cite{zhu2024tmp,zhuang2025tsinet,liu2026dsca}. TMP, for example, extracts temporal motion features and spatial image features in separate branches and then performs cross-domain fusion~\cite{zhu2024tmp}. These studies show that motion extraction and motion--appearance integration are two closely related parts of multi-frame detection. Therefore, besides obtaining reliable motion cues, an effective way to combine them with current-frame appearance features is also needed.

To address these two issues, we propose Motion Integration DETR (MI-DETR), a three-stage framework that explicitly models motion and fuses it with appearance features. The framework follows a separation--interaction--detection process. Its organization is inspired by the parallel processing of visual information in the human visual system~\cite{nassi2009parallel}. We use this principle as a high-level design motivation rather than a direct neurophysiological model.
First, to suppress background clutter while preserving target-related motion cues, Recurrent Interpretable Motion Cue Aggregation (RIMCA) maintains a recurrent temporal state that accumulates motion across consecutive frames. It produces a causal and spatially aligned motion representation. The current infrared frame is processed through a separate pathway to provide appearance features. Second, Pathway Mutual Interaction (PMI) preserves separate appearance and motion pathways while enabling bidirectional, query-guided feature exchange between their multi-scale features. This interaction combines motion cues with current-frame appearance features before detection. Finally, an RT-DETR-based detector~\cite{rtdetr} uses the refined features for end-to-end target localization. Experiments on DAUB-R, ITSDT-15K, and IRDST-H show that explicit motion modeling and pathway interaction improve moving infrared small target detection. MI-DETR also maintains real-time inference efficiency, reaching 34.60 FPS under the unified single-RTX~3090 protocol. Fig.~\ref{fig:accuracy_comparison} illustrates this accuracy--efficiency trade-off on IRDST-H.

Our main contributions are summarized as follows:
\begin{itemize}

\item We characterize moving ISTD from the relationship between target motion and background dynamics. Dynamic background changes can introduce motion cues that interfere with target-related motion information, making reliable motion extraction difficult in complex scenes.

\item We propose Motion Integration DETR (MI-DETR), a three-stage separation--interaction--detection framework for moving ISTD. MI-DETR contains two key components. RIMCA suppresses background clutter while accumulating target-related motion cues to construct an explicit motion representation. PMI then preserves separate appearance and motion pathways while enabling bidirectional, query-guided feature exchange, addressing the integration of motion cues with current-frame appearance features.

\item Extensive experiments on DAUB-R, ITSDT-15K, and IRDST-H demonstrate the effectiveness of MI-DETR. The proposed method achieves state-of-the-art performance across the three moving infrared small-target benchmarks while maintaining real-time inference efficiency.

\end{itemize}
\section{Related Work}
\label{related_work}

\subsection{Single-Frame Infrared Small Target Detection}

Single-frame ISTD detects small targets from a single infrared image. Early model-driven methods rely on hand-crafted features and predefined priors. Morphological filtering~\cite{tom1993morphology} removes background structures with designed operators. Local contrast methods~\cite{chen2013lcm} enhance targets by comparing them with nearby regions. Low-rank and sparse decomposition~\cite{gao2013infrared} separates sparse targets from structured backgrounds. These methods are interpretable and do not require training data. However, their performance often depends on assumptions about the target and background. These assumptions may be less reliable in complex scenes.

Data-driven methods learn target representations directly from infrared images. ACM~\cite{dai2021asymmetric} improves contextual feature interaction, DNA-Net~\cite{DNAnetli2022dense} preserves small-target features through dense nested connections, ISNet~\cite{zhang2022isnet} introduces shape information, and UIU-Net~\cite{wu2022uiu} strengthens multi-scale feature representation. Transformer-based methods further improve long-range contextual modeling under complex backgrounds~\cite{liu2023transformer}. These methods improve spatial feature learning and reduce the dependence on hand-crafted priors. They are effective when the target can be separated from the background by appearance cues. However, a single frame does not contain target motion or temporal changes across frames. This limitation is important for weak moving targets whose appearance may be close to the surrounding background. It motivates the use of temporal information in multi-frame ISTD.

\subsection{Multi-Frame Infrared Small Target Detection}

Multi-frame ISTD uses consecutive infrared frames to exploit temporal information and capture motion cues. Early deep methods mainly focus on spatio-temporal feature learning. Spatial--temporal frameworks combine appearance information with temporal features~\cite{du2021spatial}. STDMANet~\cite{yan2023stdmanet}, DTUM~\cite{li2023directiondtum}, and SSTNet~\cite{chen2024sstnet} further model changes across frames through differential features, direction-aware temporal modeling, or recurrent processing. These studies show that temporal information can provide cues that are not available in a single image.

Recent methods pay more attention to motion modeling under complex backgrounds. Tridos~\cite{Duan_2024_Tridos} introduces frequency-domain information to improve temporal representation. LMAFormer~\cite{huang2024lmaformer} focuses on local motion and reduces interference from moving backgrounds. MOCID~\cite{zhang2025mocid} models motion context and frame-to-frame displacement differences between targets and backgrounds. STME~\cite{peng2025moving} uses mixed spatio-temporal encoding. MoPKL and iMoPKL~\cite{chen2025motion,iMoPKL} use motion prior knowledge to improve moving-target detection. OSFormer~\cite{qin2025osformer} uses adaptive filtering for infrared video small object detection. TDCNet~\cite{fang2026tdcnet} models temporal differences and considers pseudo-motion clutter. These methods use different forms of temporal information, but they share the same challenge: background changes can also produce strong motion responses. Reliable target-related motion information is therefore still important in moving ISTD.

Motion information must also be combined with the appearance of the current frame. TMP~\cite{zhu2024tmp} uses separate temporal and spatial branches and then performs cross-domain fusion. TSI-Net~\cite{zhuang2025tsinet} introduces temporal--semantic feature interaction. DSCA-Net~\cite{liu2026dsca} uses separate spatial and temporal streams with cross-attention fusion. These methods show that motion extraction and motion--appearance interaction are closely related. A useful temporal representation should first provide reliable motion cues. The detector should then combine these cues with spatial appearance information. This separation is also useful for analysis because the quality of the motion input and the effect of feature interaction can be evaluated independently.

MI-DETR follows this direction but makes these two parts explicit. RIMCA first constructs a causal motion representation from consecutive infrared frames. It reduces background responses and keeps useful temporal responses. The motion representation and the current frame are then encoded by separate pathways. PMI performs bidirectional interaction between these pathways before detection. This design directly matches the two problems described in the Introduction: reliable motion representation and effective motion--appearance interaction.

\subsection{DETR-Based Object Detection}

DETR~\cite{carion2020endtoendobjectdetectiontransformers} formulates object detection as an end-to-end set prediction problem. It removes the need for many hand-designed post-processing steps used in conventional detectors. Deformable DETR~\cite{zhu2021deformabledetrdeformabletransformers} improves convergence and computation through sparse deformable attention. RT-DETR~\cite{rtdetr} further develops the DETR framework for real-time detection. These detectors mainly focus on the final object-detection stage. In MI-DETR, RT-DETR is used only for final target classification and localization. RIMCA and PMI operate before the detector. They provide the explicit motion representation and motion--appearance interaction used by the final detection stage.

\section{Methods}
\label{sec:method}

\subsection{Architecture Overview}
\label{sec:overview}


Human visual processing can be broadly divided into low-level feature processing, intermediate visual processing, and high-level object recognition~\cite{kandel2021principles}. MI-DETR adopts a similar hierarchical organization, progressing from motion representation to motion--appearance interaction and finally to target detection. In addition, its dual-pathway formulation resembles the parallel processing of visual information in the human visual system~\cite{nassi2009parallel}.

\begin{figure*}[!t]
\centering
\includegraphics[width=0.99\linewidth]{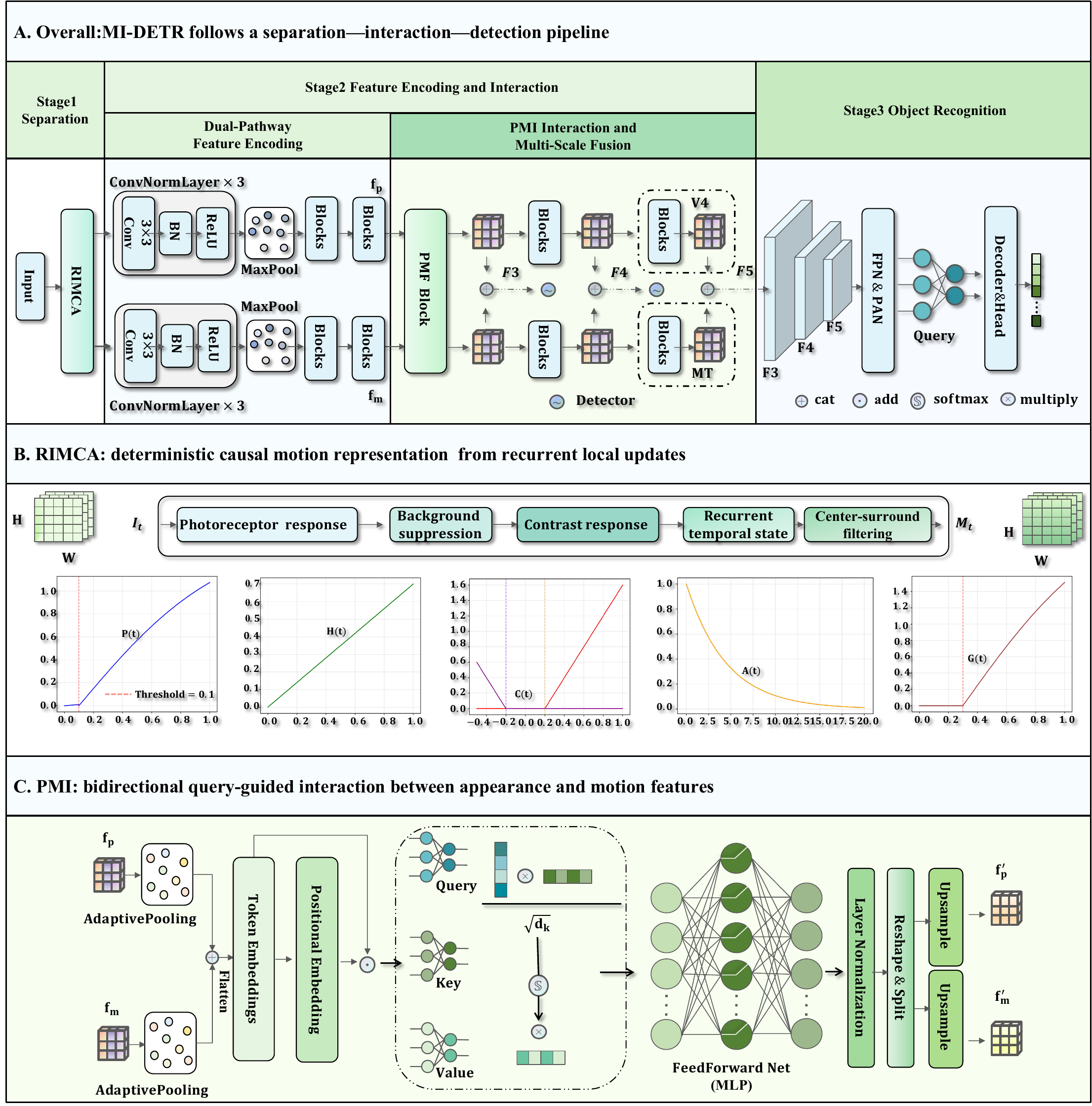}
\caption{Overall architecture of MI-DETR. RIMCA constructs the motion representation, PMI performs bidirectional interaction between the appearance and motion pathways, and RT-DETR performs the final target classification and localization.}
\label{fig:framework}
\end{figure*}

Following this organization, Fig.~\ref{fig:framework}(a) presents the overall architecture of MI-DETR. The motion pathway first constructs an explicit motion representation through the RIMCA module, as illustrated in Fig.~\ref{fig:framework}(b). The resulting motion cues are then integrated with appearance features through PMI, as shown in Fig.~\ref{fig:framework}(c), after which the refined features are fed into the detection head for target classification and localization.

For frame $I_t$, RIMCA produces a causal motion representation $M_t$ that is aligned with the current frame. Two ResNet-18 pathways~\cite{he2015deepresiduallearningimage} encode $I_t$ and $M_t$ separately. We denote the three feature levels used for detection as $F_3$, $F_4$, and $F_5$, from high to low spatial resolution. PMI first interacts the appearance and motion features at $F_3$. The refined $F_3$ features then pass through deeper backbone blocks to obtain $F_4$ and $F_5$. At each level, the corresponding appearance and motion features are concatenated and sent to RT-DETR~\cite{rtdetr}. In this way, the two pathways remain separate during feature extraction but can guide each other before detection.

\subsection{Preliminary Analysis of Motion Representations}
\label{sec:motion_preliminary}

This subsection explains the design motivation of Stage~I. Two observations are important for motion representation in moving ISTD. First, moving backgrounds and relative platform motion can produce temporal responses that interfere with target motion~\cite{huang2024lmaformer,zhang2025mocid}. Second, temporal information from consecutive frames is useful for modeling moving targets~\cite{zhuang2025tsinet}. Based on these observations, we examine representative operators from two aspects: background interference and temporal information. Let $t$ denote the frame index. Let $I_t\in\mathbb{R}^{H\times W}$ be the current infrared frame, where $H$ and $W$ are the image height and width. Let $M_t^{(r)}\in\mathbb{R}^{H\times W}$ denote the candidate motion representation produced by operator $r$.

\noindent\textbf{Background interference.}
Frame differencing directly measures image changes between nearby frames. We use the two-frame and three-frame forms
\begin{equation}
\begin{aligned}
M_t^{\mathrm{D2}}&=\left|I_t-I_{t-1}\right|,\\
M_t^{\mathrm{D3}}&=
\min\!\left(
\left|I_t-I_{t-1}\right|,
\left|I_{t-1}-I_{t-2}\right|
\right).
\end{aligned}
\label{eq:d2d3_pre}
\end{equation}
following standard frame-difference formulations~\cite{jain1979difference,zhang2012threeframe}. Here $I_{t-1}$ and $I_{t-2}$ are the previous two frames. The absolute value and $\min(\cdot,\cdot)$ are applied pixel-wise. D2 and D3 measure temporal changes, but they do not distinguish target changes from background changes.

Farneb\"ack optical flow~\cite{farneback2003twoframe} estimates displacement between adjacent frames. Let $\mathbf{x}=(x,y)^{\mathrm T}$ be a pixel location and let $\mathbf d(\mathbf x)=[u(\mathbf x),v(\mathbf x)]^{\mathrm T}$ be its estimated displacement. We use the displacement magnitude as the candidate representation:
\begin{equation}
M_t^{\mathrm{Flow}}(\mathbf x)=
\lVert\mathbf d(\mathbf x)\rVert_2=
\sqrt{u(\mathbf x)^2+v(\mathbf x)^2},
\label{eq:farneback_pre}
\end{equation}
where $u(\mathbf x)$ and $v(\mathbf x)$ are the horizontal and vertical displacement components. Optical flow provides useful motion information, but background and platform motion can also produce displacement responses~\cite{huang2024lmaformer,zhang2025mocid}.

PQFT~\cite{guo2008spatio} is a spatio-temporal saliency method. It combines appearance channels with a temporal-difference channel $D_t=|I_t-I_{t-\tau}|$, where $\tau$ is the temporal lag. For grayscale infrared input, the color-opponent channels are absent. We use the available intensity and temporal terms to obtain
\begin{equation}
M_t^{\mathrm{PQFT}}=
 g*\left|
\mathcal F_Q^{-1}\left\{
\frac{Q_t}{|Q_t|+\varepsilon}
\right\}
\right|^2,
\label{eq:pqft_pre}
\end{equation}
where $Q_t$ is the quaternion Fourier transform of the input channels, $\mathcal F_Q^{-1}$ is the inverse transform, $g$ is a Gaussian smoothing kernel, $*$ denotes convolution, and $\varepsilon>0$ avoids division by zero. PQFT combines spatial and temporal information, but it is not designed specifically for infrared background suppression.

The above operators describe temporal change, displacement, or spatio-temporal saliency. However, background responses can still remain. MNWTH~\cite{bai2010newtophat} provides a different view by using local morphological processing to suppress background structures and enhance small infrared targets. MOG2~\cite{zivkovic2004mog,zivkovic2006adaptive} uses temporal history to estimate and update a background model. Although MNWTH and MOG2 were developed for different tasks, both show the value of explicit background processing. These observations suggest that background responses should be reduced before target-related temporal information is accumulated.

\noindent\textbf{Temporal information across frames.}
Background suppression alone does not retain target information from previous frames. MNWTH processes the current frame and does not use temporal history. D2, D3, Farneb\"ack flow, and PQFT use only a small number of nearby frames and do not maintain a persistent temporal state. MOG2 maintains temporal statistics, but its state is mainly used to update the background model.

For moving ISTD, temporal information is also needed to describe target motion across frames~\cite{zhuang2025tsinet,huang2024lmaformer}. A persistent state provides a simple way to carry information from previous frames to the current frame. The Devillard--Heit cellular-automaton (CA) retina~\cite{devillard2014retina} provides a stateful example. We summarize its temporal update as
\begin{equation}
S_t^{\mathrm{DH}}=
\Phi_{\mathrm{DH}}\left(I_t,S_{t-1}^{\mathrm{DH}}\right),
\label{eq:dh_state}
\end{equation}
where $S_t^{\mathrm{DH}}$ and $S_{t-1}^{\mathrm{DH}}$ denote the current and previous states, and $\Phi_{\mathrm{DH}}$ denotes the state-update rule. This notation shows that the current state depends on the current input and the previous state. However, the Devillard--Heit model is not designed specifically for target--background separation in moving ISTD.

\noindent\textbf{Motivation for RIMCA.}
The above analysis raises a natural question: can we design a motion representation that reduces background interference while keeping useful temporal information? To address this question, we design RIMCA. It first suppresses local background responses to reduce background clutter. It then computes the change between consecutive processed frames to capture temporal responses. Finally, a recurrent state accumulates these responses over time to retain information from previous frames. In this way, RIMCA combines background suppression and temporal information accumulation in one motion representation. The controlled experiments in Sec.~\ref{subsec:motion_detector_exp} further verify the effectiveness of this design.

\subsection{Stage~I: Explicit Motion Representation with RIMCA}
\label{sec:rimca_detail}

RIMCA follows three steps. It first reduces local background responses. It then measures the temporal change between consecutive processed frames. Finally, it keeps these temporal responses in a recurrent state and reads out the final motion representation.

\noindent\textbf{Spatial background suppression.}
The first step produces a background-suppressed spatial response before the temporal update. For normalized input $I_t$, RIMCA applies an intensity-adaptation transform
\begin{equation}
P_t=\mathrm{Adapt}\left(I_t;\theta_p,g_p\right),
\label{eq:rimca_photo}
\end{equation}
where, for a normalized pixel value $x$,
\begin{equation}
\mathrm{Adapt}(x;\theta_p,g_p)=
\begin{cases}
g_p\tanh(x-\theta_p), & x>\theta_p,\\
0.1x, & x\leq\theta_p.
\end{cases}
\label{eq:rimca_adapt}
\end{equation}
Here $P_t$ is the adapted intensity response. The parameters $\theta_p$ and $g_p$ denote the intensity threshold and gain, respectively. RIMCA then subtracts a local Gaussian response to reduce slowly varying background components:
\begin{equation}
H_t=\left[P_t-\sigma_h(K_h*P_t)\right]_{+},
\label{eq:rimca_inhibition}
\end{equation}
followed by a contrast response
\begin{equation}
C_t=\left[g_b(H_t-\theta_b)\right]_{+}.
\label{eq:rimca_contrast}
\end{equation}
Here $H_t$ is the background-suppressed response and $C_t$ is the resulting contrast response. The symbol $*$ denotes convolution, and $[z]_{+}=\max(z,0)$. The kernel $K_h$ is a normalized $3\times3$ Gaussian kernel with width $\sigma_G$. The parameters $\sigma_h$, $g_b$, and $\theta_b$ control inhibition strength, contrast gain, and contrast threshold, respectively.

\noindent\textbf{Recurrent temporal accumulation.}
The second step measures temporal change on the processed response $C_t$. For $t>1$, RIMCA uses the absolute difference between two consecutive responses. The first frame has no previous response, so it is initialized with the Sobel gradient magnitude:
\begin{equation}
R_t=
\begin{cases}
\beta\lVert\nabla C_t\rVert, & t=1,\\
\beta\left|C_t-C_{t-1}\right|, & t>1,
\end{cases}
\label{eq:rimca_difference}
\end{equation}
where $\nabla C_t$ is the Sobel spatial gradient of $C_t$, and $\lVert\nabla C_t\rVert$ is its pixel-wise magnitude. For $t>1$, $R_t$ is the frame-to-frame temporal response; for $t=1$, it is the initialization used because no previous processed frame is available. The absolute difference is computed pixel-wise, and $\beta$ scales $R_t$. RIMCA then updates its temporal state using
\begin{equation}
A_t=\alpha A_{t-1}+(1-\alpha)R_t,
\label{eq:rimca_recurrent}
\end{equation}
where $A_t$ is the recurrent state, $A_{t-1}$ is the previous state, and $\alpha\in[0,1)$ controls temporal retention. For example, responses from recent frames contribute to the current state, while older responses gradually decay. All RIMCA states are reset at the start of each sequence.

\noindent\textbf{Motion-representation readout.}
The final step combines the current spatial response with the recurrent state:
\begin{equation}
U_t=C_t+\gamma_aA_t,
\label{eq:rimca_integration}
\end{equation}
where $U_t$ combines the current contrast response $C_t$ with the recurrent state $A_t$, and $\gamma_a$ controls the contribution of $A_t$. A center--surround filter then emphasizes compact local responses by subtracting a wider Gaussian response from a narrower one:
\begin{equation}
\begin{aligned}
K_m&=G_{\sigma_1}-w_{\mathrm{surr}}G_{\sigma_2},\\
M_t^{\mathrm{spa}}&=K_m*U_t,
\qquad
M_t^{\mathrm{tem}}=\gamma_{\tau}A_t.
\end{aligned}
\label{eq:rimca_dog}
\end{equation}
Here $K_m$ is the center--surround kernel. The terms $G_{\sigma_1}$ and $G_{\sigma_2}$ are Gaussian kernels with widths $\sigma_1$ and $\sigma_2$, respectively, and $w_{\mathrm{surr}}$ weights the wider surround kernel. The response $M_t^{\mathrm{spa}}$ is the center--surround spatial response, while $M_t^{\mathrm{tem}}$ is the recurrent temporal response scaled by $\gamma_{\tau}$. The two terms are combined as
\begin{equation}
G_t=g_m\left[\tanh\left(M_t^{\mathrm{spa}}+M_t^{\mathrm{tem}}-\theta_m\right)\right]_{+},
\label{eq:rimca_ganglion}
\end{equation}
where $G_t$ is the filtered readout response, $g_m$ denotes the gain coefficient, and $\theta_m$ is its activation threshold. The final motion representation is
\begin{equation}
M_t=\mathrm{Enhance}\left(\eta_mG_t+(1-\eta_m)A_t\right),
\label{eq:rimca_output}
\end{equation}
where $\eta_m$ balances the filtered response $G_t$ and recurrent state $A_t$. The spatial support of $K_m$ is denoted by $k_m$. The final enhancement step applies a power transform, bilateral filtering, and normalization:
\begin{equation}
\mathrm{Enhance}(X)=255\,\frac{\mathcal{B}([X]_{+}^{\gamma_p})}{\max_{\mathbf{x}}\mathcal{B}([X]_{+}^{\gamma_p})(\mathbf{x})+\epsilon},
\label{eq:rimca_enhance}
\end{equation}
where $X$ is the input response to the enhancement step, $\mathcal B$ denotes bilateral filtering~\cite{tomasi1998bilateral}, and $\gamma_p$ is the power exponent. The coordinate $\mathbf x$ indexes image pixels, so $\max_{\mathbf x}$ is the maximum filtered value over the image. The constant $\epsilon>0$ avoids division by zero, and the factor 255 rescales the normalized output.

We analyze six RIMCA parameters in Table~\ref{tab:rimca_parameter_sensitivity}: temporal retention $\alpha$, temporal-response scale $\beta$, activation threshold $\theta_m$, readout weight $\eta_m$, center--surround support $k_m$, and Gaussian width $\sigma_G$. Their numerical values are reported in the sensitivity analysis. The remaining gains, thresholds, and filtering constants use one shared configuration across all datasets and experiments.

\begin{figure}[!t]
\centering
\includegraphics[width=0.99\linewidth]{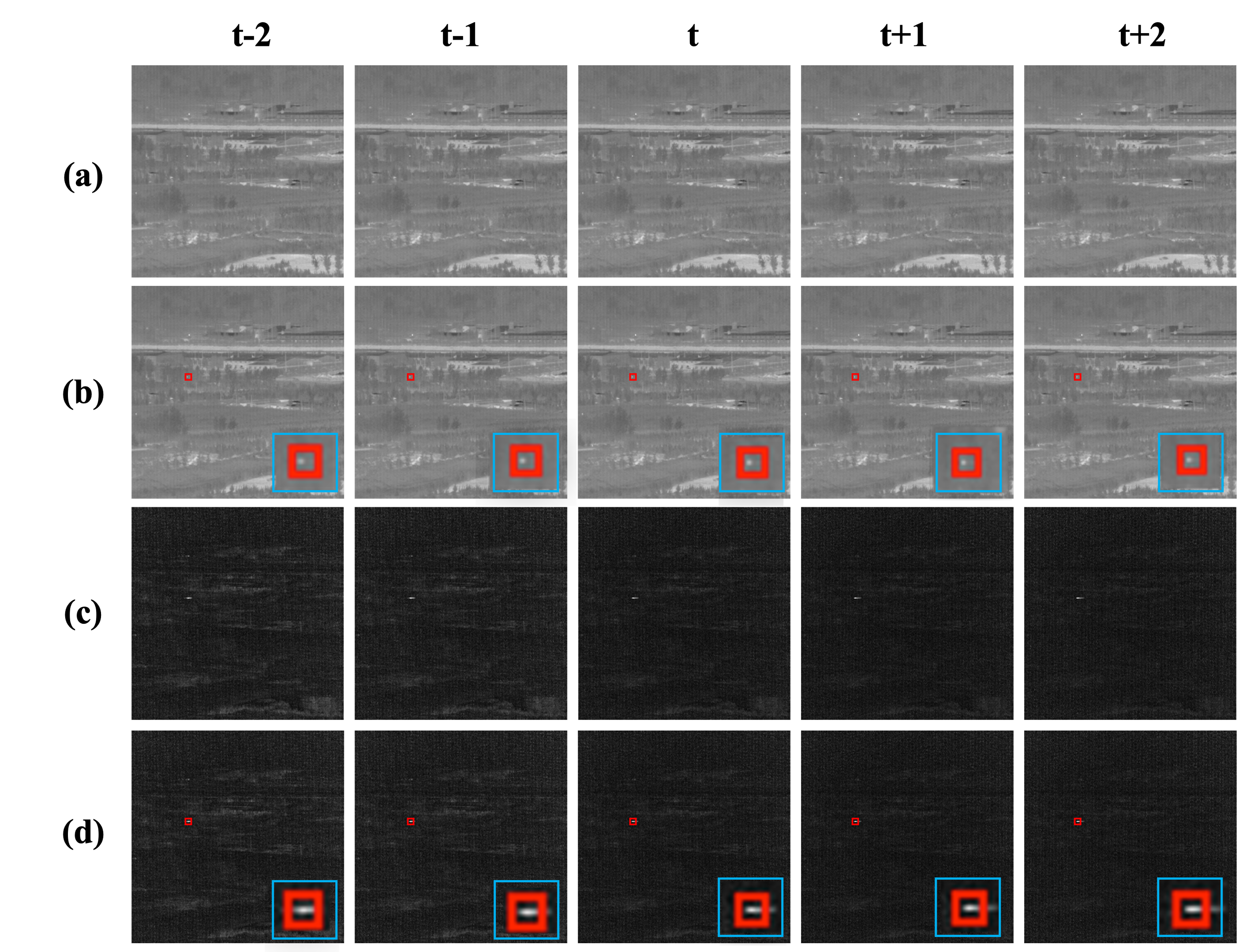}
\caption{Qualitative visualization of RIMCA over five consecutive infrared frames. (a, b) Appearance inputs without and with ground-truth (GT) annotations. (c, d) Corresponding RIMCA motion representations without and with GT annotations. Red boxes indicate GT targets, and blue boxes denote zoom-in regions.}
\label{fig:rimca_motion_extraction}
\end{figure}

RIMCA processes frames in temporal order. Each $M_t$ uses only the current frame and states from earlier frames, so no future frame is used. The states are reset at every sequence boundary. Fig.~\ref{fig:rimca_motion_extraction} shows examples of the resulting motion representations.

\subsection{Stage~II: Motion--Appearance Interaction with PMI}
\label{sec:pmi_encoder}

Stage~I provides the motion representation $M_t$, while $I_t$ provides the current-frame appearance information. PMI keeps the two pathways separate and lets each pathway guide the other. The interaction is first applied at the $F_3$ level. We use $s=\mathrm P$ for the appearance pathway and $s=\mathrm M$ for the motion pathway. If $B$ is the batch size, $C$ is the channel number, and $H_3\times W_3$ is the spatial size, the two features have the form
\begin{equation}
F_3^{s}\in\mathbb{R}^{B\times C\times H_3\times W_3},
\qquad s\in\{\mathrm P,\mathrm M\}.
\label{eq:pmi_input_shape}
\end{equation}

\noindent\textbf{Compact token representation.}
Full attention over all $H_3W_3$ positions is expensive. PMI therefore compresses each feature to a smaller $H_a\times W_a$ grid, where $H_a$ and $W_a$ are the pooled height and width. It combines adaptive average pooling and adaptive max pooling:
\begin{equation}
\begin{aligned}
\bar F^{s}&=\omega_{\mathrm{avg}}^{s}P_{\mathrm{avg}}(F_3^{s})
+\omega_{\mathrm{max}}^{s}P_{\mathrm{max}}(F_3^{s}),\\
X^{s}&=\mathrm{Flatten}(\bar F^{s})+E^{s}.
\end{aligned}
\label{eq:pmi_tokens}
\end{equation}
Here $P_{\mathrm{avg}}$ and $P_{\mathrm{max}}$ are adaptive average- and max-pooling operators. The terms $\omega_{\mathrm{avg}}^{s}$ and $\omega_{\mathrm{max}}^{s}$ are learnable pooling weights, and $\bar F^s$ is the pooled feature. The operator $\mathrm{Flatten}$ converts the pooled grid into a token sequence, while $E^s$ is a positional embedding. Thus, $X^s\in\mathbb{R}^{B\times N_a\times C}$ is the token sequence with $N_a=H_aW_a$ tokens.

\noindent\textbf{One-way cross-attention.}
Consider one direction first. Here $a$ denotes the pathway being refined and $b$ denotes the pathway that provides guidance. PMI uses pathway $b$ as the query source, while pathway $a$ provides the keys and values. The learned linear projections give $Q_b=X^bW_b^Q$, $K_a=X^aW_a^K$, and $V_a=X^aW_a^V$. With $h$ attention heads and per-head dimension $d=C/h$,
\begin{equation}
A_{b\rightarrow a}=\mathrm{Softmax}\left(\frac{Q_bK_a^{\mathrm T}}{\sqrt d}\right),
\qquad
Z_a=W_o^a\left(A_{b\rightarrow a}V_a\right).
\label{eq:pmi_guided_attention}
\end{equation}
Here $Q_b$, $K_a$, and $V_a$ are the query, key, and value tokens. The matrix $A_{b\rightarrow a}$ contains the attention weights from pathway $b$ to pathway $a$, and $Z_a$ is the guided response for pathway $a$. The matrices $W^Q$, $W^K$, $W^V$, and $W_o$ are learned linear projections. For example, when motion guides appearance, the motion query determines which appearance tokens receive more attention, but the content being combined still comes from the appearance pathway. Thus, one pathway guides the other without replacing its features.

The guided response is added to the original tokens and then processed by layer normalization (LN) and a feed-forward network (FFN):
\begin{equation}
U_a=\lambda_1X^a+\lambda_2Z_a,
\qquad
Y_a=\lambda_3U_a+\lambda_4\mathrm{FFN}_a\left(\mathrm{LN}(U_a)\right).
\label{eq:pmi_residual}
\end{equation}
Here $U_a$ is the first residual fusion and $Y_a$ is the refined token sequence. The coefficients $\lambda_1$--$\lambda_4$ are learnable residual weights. The refined tokens $Y_a$ are reshaped to the compact grid, resized to the original $F_3$ spatial size, and added back to pathway $a$. We denote this one-way operation by $\Phi(F_a,F_b)$, where the first argument is the pathway being refined and the second is the guiding pathway.

\noindent\textbf{Bidirectional cross-attention and multi-scale propagation.}
PMI applies the same idea in both directions with independent parameters:
\begin{equation}
\widehat F_3^{\mathrm P}=\Phi_{\mathrm P}(F_3^{\mathrm P},F_3^{\mathrm M}),
\qquad
\widehat F_3^{\mathrm M}=\Phi_{\mathrm M}(F_3^{\mathrm M},F_3^{\mathrm P}).
\label{eq:pmi_bidirectional}
\end{equation}
Thus, motion guides appearance refinement and appearance guides motion refinement. Let $\mathcal B_4^s$ and $\mathcal B_5^s$ denote the next two backbone stages. The refined $F_3$ features pass through them as
\begin{equation}
\widehat F_4^{s}=\mathcal{B}_4^{s}(\widehat F_3^{s}),
\qquad
\widehat F_5^{s}=\mathcal{B}_5^{s}(\widehat F_4^{s}),
\qquad s\in\{\mathrm P,\mathrm M\}.
\label{eq:pmi_hierarchical}
\end{equation}
At each level $l$, we concatenate the refined appearance feature $\widehat F_l^{\mathrm P}$ and motion feature $\widehat F_l^{\mathrm M}$ along the channel dimension, denoted by $\Vert$:
\begin{equation}
F_l=\widehat F_l^{\mathrm P}\Vert\widehat F_l^{\mathrm M},
\qquad l\in\{3,4,5\}.
\label{eq:pmi_multiscale}
\end{equation} The resulting multi-scale features $\{F_3,F_4,F_5\}$ are passed to the detector. The same compact-grid size and attention settings are used for all experiments and are given in Sec.~\ref{sec:implementation_details}.

\subsection{Stage~III: Detection with RT-DETR}
\label{sec:decoder}

Stage~III sends the multi-scale features $\{F_3,F_4,F_5\}$ to the standard RT-DETR encoder and decoder~\cite{rtdetr}. RT-DETR predicts the target class and bounding box. No new temporal modeling is added in this stage. Thus, RIMCA constructs the motion representation, PMI combines motion and appearance features, and RT-DETR produces the final detections.

\section{Experiments}
\label{Sec:EXP}
The experiments follow the Methods structure. We first compare different motion representations under the same detector to evaluate the Stage~I representation. We then compare the complete MI-DETR with state-of-the-art methods. Finally, controlled ablations analyze the RIMCA readout, RIMCA parameter sensitivity, PMI, and the complete dual-pathway configuration across different detectors.

\subsection{Datasets and Evaluation Metrics}
We evaluate MI-DETR on three benchmark datasets for moving infrared small target detection: \textit{ITSDT-15K}~\cite{itsdt15k}, \textit{IRDST-H}~\cite{iMoPKL,RDIAN}, and \textit{DAUB-R}~\cite{iMoPKL,720626420933459968}. We follow the data splits reported in iMoPKL~\cite{iMoPKL} so that the benchmark results are directly comparable with prior moving infrared small-target methods. Table~\ref{tab:datasets} summarizes the dataset statistics.
\begin{table}[!t]
\centering
\caption{Summary of datasets used in our experiments. Train, Val, and Test denote the numbers of training, validation, and test samples, respectively; -- indicates that no separate validation split is used.}
\label{tab:datasets}
\TableFont
\setlength{\tabcolsep}{7.0pt}
\renewcommand{\arraystretch}{1.05}
\begin{tabular}{lcccc}
\toprule
\textbf{Dataset} & \textbf{Train} & \textbf{Val} & \textbf{Test} & \textbf{Total} \\
\midrule
\textit{ITSDT-15K}~\cite{itsdt15k}     & 10{,}000 & --      & 5{,}000 & 15{,}000 \\
\textit{IRDST-H}~\cite{iMoPKL} & 8{,}725  & 2{,}694 & 5{,}354 & 16{,}773 \\
\textit{DAUB-R}~\cite{iMoPKL}  & 7{,}500  & 2{,}000 & 4{,}277 & 13{,}777 \\
\bottomrule
\end{tabular}
\end{table}

\noindent\textbf{Evaluation metrics.}
Following standard object-detection evaluation~\cite{lin2014microsoft}, we report precision $P=\mathrm{TP}/(\mathrm{TP}+\mathrm{FP})$, recall $R=\mathrm{TP}/(\mathrm{TP}+\mathrm{FN})$, and $F1=2PR/(P+R)$, where TP, FP, and FN denote true positives, false positives, and false negatives. mAP$_{50}$ is mean Average Precision at an intersection-over-union (IoU) threshold of 0.5 over confidence-ranked detections. For P, R, and F1, the shared evaluator selects the operating point at the maximum of the smoothed F1 curve and reports the corresponding precision and recall. The same evaluation procedure is used for all detector tables.


\subsection{Implementation Details}
\label{sec:implementation_details}
MI-DETR training is conducted on a high-performance computing cluster with NVIDIA A100 GPUs (40\,GB). The paired appearance frames and motion representations are directly resized to $512\times512$ by the RT-DETR dataset pipeline, without aspect-ratio-preserving letterbox padding.

\noindent\textbf{Training setup.}
MI-DETR is trained for 600 epochs with a batch size of 32 at an input resolution of $512\times512$. We use AdamW~\cite{loshchilov2017adamw} with an initial learning rate of $8\times10^{-5}$, weight decay of $8\times10^{-4}$, and the first-moment coefficient $\beta_1=0.937$. A 15-epoch linear learning-rate warm-up is used, while $\beta_1$ remains fixed throughout training. Appearance frames, precomputed RIMCA representations, and bounding boxes are treated as paired samples so that every geometric transformation is applied synchronously to both pathways and the annotations, preserving the pixel correspondence required by PMI.

\noindent\textbf{Inference and complexity measurement.}
At test time, the paired inputs are directly resized to $512\times512$ by the same dataset pipeline. 
The confidence threshold of 0.5 and NMS threshold of 0.3 in Fig.~\ref{fig:visual_comparison} are used only for visualization and are not used to compute the reported numerical results. The giga floating-point operations (GFLOPs), parameter counts, and frames per second (FPS) reported in Sec.~\ref{Section:experimentComplexity} retain the unified single-RTX~3090 comparison protocol. For MI-DETR, the reported throughput is the complete end-to-end RIMCA + detector pipeline, including RIMCA motion-representation construction, paired input processing, and detector inference, and reaches 34.60 FPS. The component-level RIMCA latency in the layer-wise analysis is reported separately on an A100.

\subsection{Detection-Oriented Motion Analysis}
\label{subsec:motion_detector_exp}

Table~\ref{tab:motion_detector_replacement} compares the candidate motion representations discussed in Sec.~\ref{sec:motion_preliminary} under the same detector. We keep the ResNet-18 dual-pathway detector, PMI, RT-DETR decoder, training setup, and evaluation protocol fixed. Only the motion representation is changed. The comparison includes D2, D3, Farneb\"ack flow, MNWTH, MOG2, PQFT, and the Devillard--Heit representation. A zero representation is used as the control.

\begin{table}[!t]
\centering
\caption{Detection-oriented comparison of representative motion representations on DAUB-R under a fixed detector. Only the motion representation is changed, and all variants are reported with the common mAP$_{50}$/P/R/F1 convention. Full RIMCA is the Stage~I representation used by the complete MI-DETR model; its DAUB-R detector metrics are therefore identical to the MI-DETR entry in Table~\ref{tab:quantitative_v1}.}
\label{tab:motion_detector_replacement}
\TableFont
\setlength{\tabcolsep}{4.0pt}
\renewcommand{\arraystretch}{1.08}
\begin{tabular}{lcccc}
\toprule
\textbf{Motion Representation} & \textbf{mAP$_{50}$} $\uparrow$ & \textbf{P} $\uparrow$ & \textbf{R} $\uparrow$ & \textbf{F1} $\uparrow$ \\
\midrule
Zero representation & 87.55 & 85.83 & 78.32 & 81.90 \\
Two-frame Difference~\cite{jain1979difference} & 93.94 & 90.59 & 86.54 & 88.52 \\
Three-frame Difference~\cite{zhang2012threeframe} & 90.11 & 86.12 & 81.45 & 83.72 \\
Farneb\"ack~\cite{farneback2003twoframe} & 94.07 & \textbf{94.57} & 85.32 & 89.71 \\
MNWTH~\cite{bai2010newtophat} & 88.59 & 92.97 & 77.47 & 84.52 \\
MOG2~\cite{zivkovic2004mog,zivkovic2006adaptive} & 84.36 & 83.84 & 74.79 & 79.06 \\
PQFT~\cite{guo2008spatio} & 90.15 & 88.79 & 81.24 & 84.85 \\
Devillard--Heit CA Retina~\cite{devillard2014retina} & 92.75 & 90.04 & 88.75 & 89.39 \\
\midrule
\rowcolor{gray!8}
\textbf{Full RIMCA (Ours)} & \textbf{98.00} & 93.80 & \textbf{94.90} & \textbf{94.35} \\
\bottomrule
\end{tabular}
\end{table}

\begin{table*}[!t]
\centering
\caption{Comparison with representative methods on three moving infrared small-target benchmarks. Best and second-best results are shown in \textbf{bold} and \underline{underlined}, respectively. Numeric temporal-context entries denote the number of input frames for single-frame and fixed-window methods; MI-DETR instead maintains a causal recurrent state whose historical contributions decay exponentially rather than being truncated to a fixed input window.}
\label{tab:quantitative_v1}
\TableFont
\setlength{\tabcolsep}{2.7pt}
\renewcommand{\arraystretch}{1.08}
\begin{tabular}{lcccccccccccccc}
\toprule
\multirow{2}{*}{\textbf{Method}} &
\multirow{2}{*}{\textbf{Temporal Context}} &
\multirow{2}{*}{\textbf{Venue}} &
\multicolumn{4}{c}{\textbf{DAUB-R}} &
\multicolumn{4}{c}{\textbf{ITSDT-15K}} &
\multicolumn{4}{c}{\textbf{IRDST-H}} \\
\cmidrule(lr){4-7}\cmidrule(lr){8-11}\cmidrule(lr){12-15}
& & & mAP$_{50}$ & P & R & F1 & mAP$_{50}$ & P & R & F1 & mAP$_{50}$ & P & R & F1 \\
\midrule
\multicolumn{15}{l}{\textit{Single-frame methods}} \\
\cmidrule(lr){1-15}
SANet~\cite{zhu2023SAnet} & 1 & ICASSP 2023 & 83.64 & 91.62 & 92.26 & 91.94 & 62.17 & 87.78 & 71.23 & 78.64 & 33.02 & 51.86 & 64.50 & 57.49 \\
AGPCNet~\cite{zhang2021agpcnet} & 1 & IEEE TAES 2023 & 81.25 & 84.44 & \underline{97.66} & 90.57 & 67.27 & 91.19 & 74.77 & 82.16 & 29.24 & 46.64 & 63.68 & 53.84 \\
RDIAN~\cite{RDIAN} & 1 & IEEE TGRS 2023 & 82.55 & 87.65 & 95.23 & 91.28 & 68.49 & 90.56 & 76.06 & 82.68 & 30.57 & 42.18 & 73.50 & 53.60 \\
DNANet~\cite{DNAnetli2022dense} & 1 & IEEE TIP 2023 & 83.65 & 88.74 & 95.18 & 91.85 & 70.46 & 88.55 & 80.73 & 84.46 & 31.07 & 51.09 & 61.09 & 55.64 \\
SIRST5K~\cite{lu2024sirst5k} & 1 & IEEE TGRS 2024 & 83.20 & 89.28 & 94.08 & 91.62 & 61.52 & 86.95 & 71.32 & 78.36 & 24.22 & 44.92 & 54.02 & 49.05 \\
MSHNet~\cite{liu2024MSH} & 1 & CVPR 2024 & 83.52 & 89.53 & 94.93 & 92.15 & 60.82 & 89.69 & 68.44 & 77.64 & 27.02 & 45.25 & 60.04 & 51.61 \\
CSViG~\cite{lin2024cs} & 1 & ESWA 2024 & 82.82 & 86.96 & 96.75 & 91.59 & 72.46 & 83.09 & 86.12 & 84.58 & 30.17 & 49.01 & 62.45 & 54.92 \\
SCTransNet~\cite{yuan2024sctransnet} & 1 & IEEE TGRS 2024 & 81.56 & 91.01 & 91.00 & 91.00 & 73.37 & 91.74 & 78.49 & 84.60 & 27.41 & 48.34 & 57.63 & 52.58 \\
RPCANet~\cite{wu2024rpcanet} & 1 & WACV 2024 & 83.03 & 86.29 & \textbf{97.71} & 91.65 & 62.28 & 81.46 & 77.10 & 79.22 & 29.17 & 45.17 & 65.18 & 53.36 \\
PConv~\cite{yang2025pinwheel} & 1 & AAAI 2025 & 84.03 & 90.98 & 93.36 & 92.15 & 61.19 & 88.93 & 69.69 & 78.14 & 33.07 & 50.45 & 66.43 & 57.35 \\
L2SKNet~\cite{wu2024saliency} & 1 & IEEE TGRS 2025 & 83.91 & 87.08 & 97.31 & 91.91 & 68.93 & 92.20 & 75.84 & 83.22 & 37.15 & 54.16 & 69.75 & 60.98 \\
\midrule
\multicolumn{15}{l}{\textit{Fixed-window multi-frame methods}} \\
\cmidrule(lr){1-15}
DTUM~\cite{li2023directiondtum} & 5 & IEEE TNNLS 2025 & 75.94 & 91.59 & 83.70 & 87.47 & 67.97 & 77.95 & 88.28 & 82.79 & 37.98 & 49.67 & \textbf{77.18} & 60.44 \\
TMP~\cite{zhu2024tmp} & 5 & ESWA 2024 & 74.58 & \textbf{99.32} & 75.80 & 85.98 & 77.73 & 92.97 & 84.74 & 88.67 & 38.93 & 63.99 & 61.73 & 62.84 \\
SSTNet~\cite{chen2024sstnet} & 5 & IEEE TGRS 2024 & 83.34 & 94.15 & 89.64 & 91.84 & 77.30 & 92.49 & 84.32 & 88.22 & 39.04 & \underline{65.12} & 60.77 & 62.87 \\
Tridos~\cite{Duan_2024_Tridos} & 5 & IEEE TGRS 2024 & 85.37 & 94.33 & 91.70 & 92.99 & 80.41 & 90.71 & \textbf{90.60} & \textbf{90.65} & 38.51 & 55.29 & 70.61 & 62.02 \\
STME~\cite{peng2025moving} & 5 & EAAI 2025 & 84.79 & 95.60 & 89.50 & 92.45 & 77.33 & 92.42 & 84.35 & 88.21 & 34.29 & 59.36 & 58.25 & 58.80 \\
MoPKL~\cite{chen2025motion} & 5 & AAAI 2025 & 85.06 & \underline{99.09} & 86.51 & 92.37 & 79.78 & \underline{93.29} & 86.80 & 89.92 & 40.66 & 59.26 & 69.68 & 64.05 \\
iMoPKL~\cite{iMoPKL} & 2 & IEEE TGRS 2025 & \underline{88.57} & 92.94 & 96.94 & \textbf{94.90} & \underline{80.67} & 92.28 & \underline{88.50} & \underline{90.35} & \underline{43.95} & 59.82 & \underline{74.48} & \underline{66.35} \\
\midrule
\multicolumn{15}{l}{\textit{Recurrent-state method}} \\
\cmidrule(lr){1-15}
\rowcolor{gray!8}
\textbf{MI-DETR (Ours)} & Recurrent state & -- & \textbf{98.00} & 93.80 & 94.90 & \underline{94.35} & \textbf{88.30} & \textbf{93.40} & 82.40 & 87.60 & \textbf{70.30} & \textbf{71.70} & 73.80 & \textbf{72.70} \\
\bottomrule
\end{tabular}
\end{table*}

Several candidate representations improve the detector over the zero representation. D2 raises mAP$_{50}$ from 87.55 to 93.94, Farneb\"ack reaches 94.07, and the Devillard--Heit representation reaches 92.75. These results show that temporal change, displacement, and stateful processing can all provide useful information. In contrast, MNWTH and MOG2 give smaller gains when used alone as the motion input.

Farneb\"ack gives the highest precision at 94.57, but its recall is 85.32. Full RIMCA reaches 93.80 precision and 94.90 recall. It obtains 98.00 mAP$_{50}$ and 94.35 F1, improving Farneb\"ack by 3.93 and 4.64 points, respectively. This result agrees with the motivation in Sec.~\ref{sec:motion_preliminary}: useful temporal change should be combined with background suppression and temporal retention. The readout and sensitivity analyses below examine the internal RIMCA responses and parameters. The Full RIMCA row uses the same complete MI-DETR configuration as the main DAUB-R result.


\subsection{Comparison with State-of-the-Art Methods}

\label{subsec:comparison_sota}
\subsubsection{Quantitative Comparison}


Table~\ref{tab:quantitative_v1} compares MI-DETR with representative single-frame and multi-frame methods. MI-DETR achieves the highest mAP$_{50}$ on all three benchmarks. Relative to the second-highest result, the margins are 9.43 points on DAUB-R, 7.63 points on ITSDT-15K, and 26.35 points on IRDST-H. The gain on all three datasets shows that the improvement is not limited to one benchmark.

The operating-point metrics show a more mixed pattern. On DAUB-R, iMoPKL has a slightly higher F1 (94.90 versus 94.35), while MI-DETR has a much higher mAP$_{50}$. On ITSDT-15K, Tridos and iMoPKL obtain higher recall, whereas MI-DETR gives the highest mAP$_{50}$ and precision. On IRDST-H, MI-DETR gives the highest mAP$_{50}$, precision, and F1. These results are consistent with the MI-DETR design: RIMCA provides explicit motion information and PMI combines it with current-frame appearance. The controlled experiments below analyze the two components separately.

\subsubsection{Qualitative Comparison}
\label{subsubsec:qualitative_comparison}

\begin{figure*}[!t]
\vspace{-5mm}  
\centering
\includegraphics[trim=0 35mm 0 20mm, clip, width=0.985\linewidth]{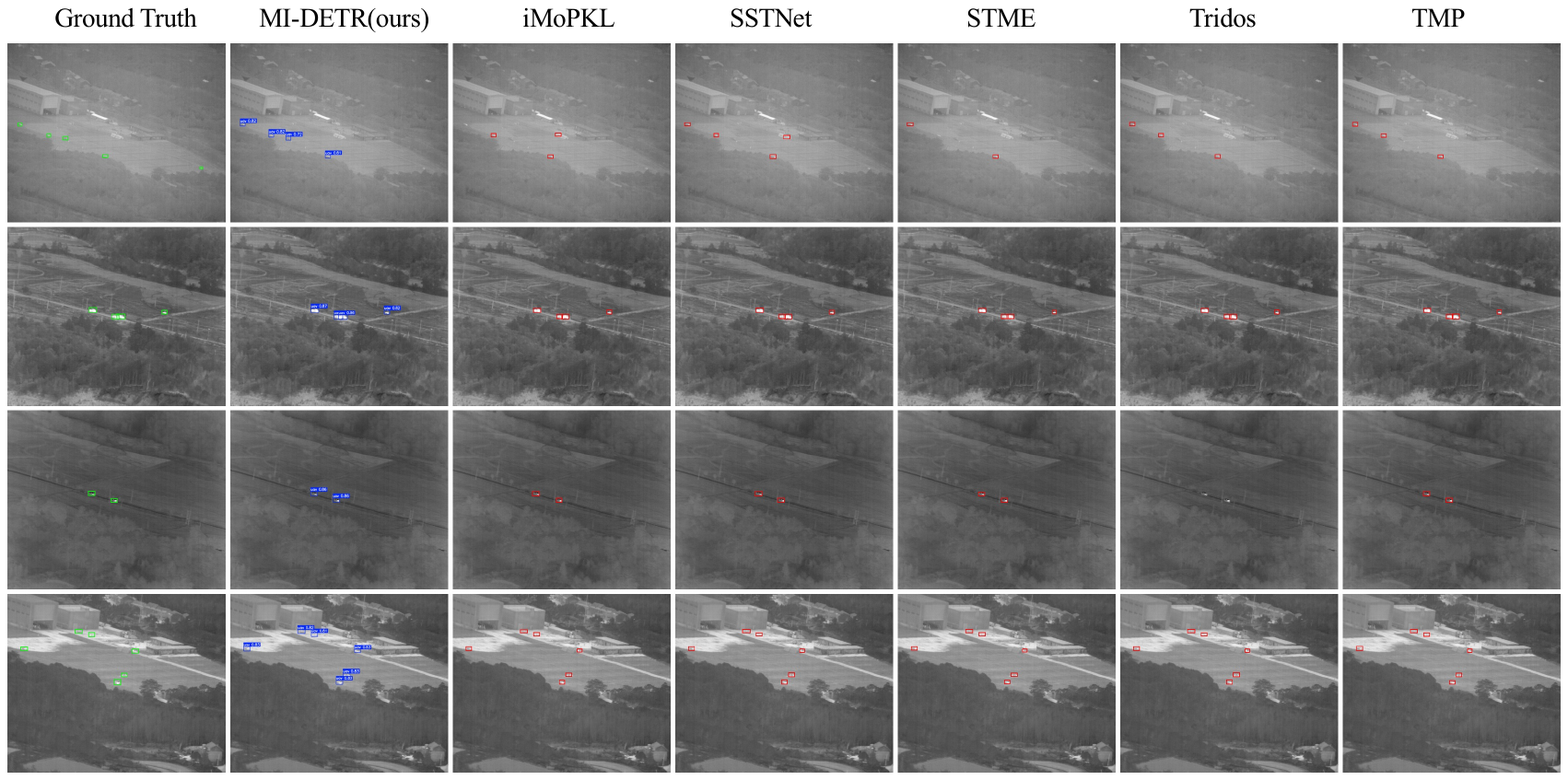}
\vspace{-2mm}  
\caption{Qualitative comparison on \textit{ITSDT-15K} (confidence: 0.5, NMS: 0.3). Green boxes denote ground truth annotations, blue boxes indicate MI-DETR predictions, and red boxes show results from five multi-frame baselines (iMoPKL~\cite{iMoPKL}, SSTNet~\cite{chen2024sstnet}, STME~\cite{peng2025moving}, Tridos~\cite{Duan_2024_Tridos}, TMP~\cite{zhu2024tmp}).}
\label{fig:visual_comparison}
\vspace{-3mm}  
\end{figure*}

To complement the aggregate metrics, Fig.~\ref{fig:visual_comparison} presents qualitative results on four scenes from \textit{ITSDT-15K}. The scenes contain small targets under different local backgrounds. In the displayed examples, MI-DETR detects the annotated targets, while some compared multi-frame baselines show missed detections or additional responses away from the target. The zoomed regions make these differences easier to inspect at the small-target scale. The examples also show why motion information alone is not sufficient. A useful detector must keep the weak target response while avoiding responses from nearby background structures. This is consistent with the motion--appearance design of MI-DETR. RIMCA provides temporal information, while PMI lets this information interact with the current-frame appearance features. These examples are used only for qualitative illustration; all quantitative conclusions are based on the complete benchmark test sets in Table~\ref{tab:quantitative_v1}.

\subsubsection{Precision-Recall Curve Analysis}
\label{subsubsec:pr_curve}

\begin{figure*}[!t]
\centering
\includegraphics[
    trim=0 0mm 0 0mm,
    clip,
    width=\textwidth
]{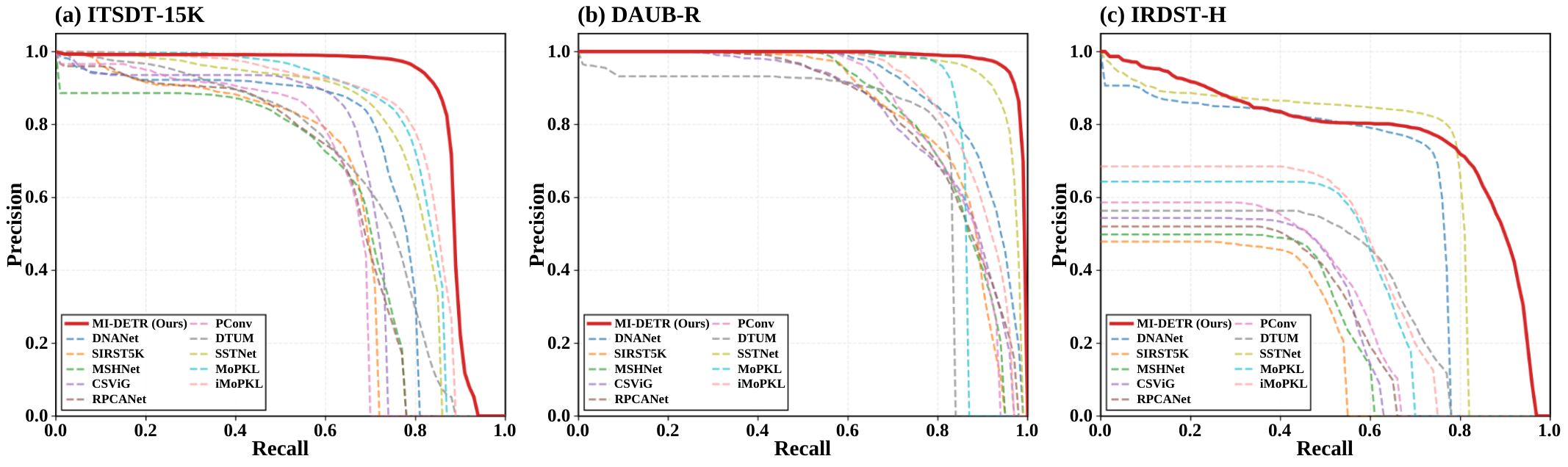}
\caption{Precision--Recall (PR) curves of MI-DETR and 11 representative methods on (a)~\textit{ITSDT-15K}, (b)~\textit{DAUB-R}, and (c)~\textit{IRDST-H}.}
\label{fig:pr}
\end{figure*}

The precision--recall (PR) curves in Fig.~\ref{fig:pr} provide the full-detector comparison corresponding to Table~\ref{tab:quantitative_v1}. On ITSDT-15K and DAUB-R, the MI-DETR curve stays at high precision over a broad recall range before dropping near the high-recall end. On IRDST-H, where the numerical gap in Table~\ref{tab:quantitative_v1} is largest, the MI-DETR curve also remains above most compared curves over much of the recall range. This behavior is consistent with the higher mAP$_{50}$ values in Table~\ref{tab:quantitative_v1}. It also shows that the advantage is not caused by one selected confidence threshold. MI-DETR keeps competitive precision while recall changes across the curve. This is important for small targets because the operating threshold can change the balance between missed targets and false responses. The PR curves describe the complete detector only. Component-level effects are evaluated by the controlled motion and interaction experiments below.

\subsubsection{Computational Complexity Analysis}
\label{Section:experimentComplexity}

\begin{table}[!t]
\centering
\caption{Accuracy--complexity comparison on IRDST-H under the unified RTX~3090 protocol. $T$ denotes temporal context: numeric entries are the numbers of input frames, while State denotes a recurrent-state method. Par. denotes millions of parameters, G denotes GFLOPs, and FPS denotes frames per second. MI-DETR FPS is measured for the complete RIMCA + detector pipeline.}
\label{tab:params_sorted_fps}
\TableFont
\setlength{\tabcolsep}{1.15pt}
\renewcommand{\arraystretch}{1.03}
\begin{tabular}{@{}lcccccccc@{}}
\toprule
\textbf{Method} & \textbf{T} & \textbf{mAP$_{50}$} & \textbf{P} & \textbf{R} & \textbf{F1} & \textbf{Par.} & \textbf{G} & \textbf{FPS} \\
\midrule

\multicolumn{9}{l}{\textit{Single-frame methods}} \\
SIRST5K~\cite{lu2024sirst5k} & 1 & 24.22 & 44.92 & 54.02 & 49.05 & 11.48 & 182.61 & 6.17 \\
DNANet~\cite{DNAnetli2022dense} & 1 & 31.07 & 51.09 & 61.09 & 55.64 & 7.22 & 135.24 & 7.21 \\
SCTransNet~\cite{yuan2024sctransnet} & 1 & 27.41 & 48.34 & 57.63 & 52.58 & 13.71 & 101.61 & 10.34 \\
RPCANet~\cite{wu2024rpcanet} & 1 & 29.17 & 45.17 & 65.18 & 53.36 & 3.21 & 382.69 & 14.81 \\
MSHNet~\cite{liu2024MSH} & 1 & 27.02 & 45.25 & 60.04 & 51.61 & 6.59 & 69.49 & 16.37 \\
AGPCNet~\cite{zhang2021agpcnet} & 1 & 29.24 & 46.64 & 63.68 & 53.84 & 14.88 & 366.15 & 18.33 \\
L2SKNet~\cite{wu2024saliency} & 1 & 37.15 & 54.16 & 69.75 & 60.98 & 3.42 & 76.00 & 30.54 \\
RDIAN~\cite{RDIAN} & 1 & 30.57 & 42.18 & 73.50 & 53.60 & \textbf{2.74} & 50.44 & 34.20 \\
PConv~\cite{yang2025pinwheel} & 1 & 33.07 & 50.45 & 66.43 & 57.35 & 2.91 & \textbf{7.89} & 40.24 \\
CSViG~\cite{lin2024cs} & 1 & 30.17 & 49.01 & 62.45 & 54.92 & 5.81 & 117.56 & \textbf{40.48} \\

\midrule
\multicolumn{9}{l}{\textit{Fixed-window multi-frame methods}} \\
MoPKL~\cite{chen2025motion} & 5 & 40.66 & 59.26 & 69.68 & 64.05 & 9.46 & 119.64 & 10.20 \\
SSTNet~\cite{chen2024sstnet} & 5 & 39.04 & 65.12 & 60.77 & 62.87 & 11.95 & 123.59 & 10.38 \\
Tridos~\cite{Duan_2024_Tridos} & 5 & 38.51 & 55.29 & 70.61 & 62.02 & 20.60 & 188.55 & 10.63 \\
TMP~\cite{zhu2024tmp} & 5 & 38.93 & 63.99 & 61.73 & 62.84 & 16.41 & 92.85 & 12.75 \\
DTUM~\cite{li2023directiondtum} & 5 & 37.98 & 49.67 & 77.18 & 60.44 & 9.64 & 128.16 & 12.77 \\
STME~\cite{peng2025moving} & 5 & 34.29 & 59.36 & 58.25 & 58.80 & 9.93 & 42.09 & 14.55 \\
iMoPKL~\cite{iMoPKL} & 2 & 43.95 & 59.82 & 74.48 & 66.35 & 34.07 & 119.13 & 28.95 \\

\midrule
\multicolumn{9}{l}{\textit{Recurrent-state method}} \\
\rowcolor{gray!8}
\textbf{MI-DETR (Ours)} &
State &
\textbf{70.30} &
\textbf{71.70} &
73.80 &
\textbf{72.70} &
32.44 &
93.90 &
34.60 \\

\bottomrule
\end{tabular}
\end{table}

Table~\ref{tab:params_sorted_fps} reports accuracy and inference speed under the unified RTX~3090 protocol. MI-DETR reaches 34.60 FPS and 70.30\% mAP$_{50}$ on IRDST-H. Among the compared multi-frame methods, it has both the highest mAP$_{50}$ and the highest FPS. The reported speed includes the complete RIMCA motion-representation construction.

For reference, iMoPKL reaches 28.95 FPS with 43.95\% mAP$_{50}$, 34.07M parameters, and 119.13 GFLOPs. MI-DETR uses 32.44M parameters and 93.90 GFLOPs. Thus, MI-DETR uses fewer parameters and GFLOPs while giving higher accuracy and throughput in this protocol. Some single-frame methods are faster, but they do not process temporal information. RIMCA uses fixed-size local filtering and pointwise recurrent updates, so its per-frame computation and state storage scale linearly with image area. The A100 component latency in Sec.~\ref{subsec:rimca_layer_ablation} remains a separate diagnostic profile.

\subsection{Ablation Study}
\label{sec:ablation}
We next analyze RIMCA and PMI separately. The RIMCA experiments study its intermediate readouts and parameter sensitivity. The PMI experiments compare single-pathway, simple aggregation, and interactive variants. All DAUB-R ablations that report the full configuration use the same model as Table~\ref{tab:quantitative_v1}.
\subsubsection{Intermediate RIMCA Readout Ablation}
\label{subsec:rimca_layer_ablation}

Table~\ref{tab:rimca_layerwise_ablation} sends different RIMCA readouts to the same detector. R0 uses a zero representation. R1--R5 use $P_t$, $H_t$, $C_t$, $A_t$, and $G_t$, respectively. R6 uses the fusion $\eta_m G_t+(1-\eta_m)A_t$ before the final enhancement step. R7 uses the final RIMCA representation $M_t$.

\begin{table}[!t]
\centering
\caption{Intermediate RIMCA readout ablation on DAUB-R. R7 is the full RIMCA output. Latency is the cumulative RIMCA processing time per frame in milliseconds; profile FPS is measured on an A100 (40\,GB). These two columns are separate from the unified RTX~3090 end-to-end result.}
\label{tab:rimca_layerwise_ablation}
\TableFont
\setlength{\tabcolsep}{1.55pt}
\renewcommand{\arraystretch}{1.04}
\begin{tabular}{@{}clcccccc@{}}
\toprule
\textbf{ID} & \textbf{Readout} & \textbf{mAP$_{50}$} & \textbf{P} & \textbf{R} & \textbf{F1} & \textbf{Latency (ms)} & \textbf{FPS} \\
\midrule
R0 & Zero representation & 87.55 & 85.83 & 78.32 & 81.90 & 0.000 & 56.36 \\
R1 & Adapted response $P_t$ & 90.25 & \textbf{95.62} & 77.08 & 85.35 & 0.150 & 55.88 \\
R2 & Suppressed $H_t$ & 86.97 & 87.99 & 78.27 & 82.84 & 0.233 & 55.63 \\
R3 & Contrast $C_t$ & 87.92 & 80.02 & 80.90 & 80.45 & 0.312 & 55.38 \\
R4 & Recurrent state $A_t$ & \underline{95.17} & \underline{94.14} & \underline{90.93} & \underline{92.50} & 0.442 & 54.99 \\
R5 & Filtered response $G_t$ & 86.65 & 85.19 & 78.07 & 81.48 & 0.765 & 54.03 \\
R6 & $G_t$--$A_t$ fusion & 92.81 & 90.16 & 84.29 & 87.13 & 0.802 & 53.92 \\
\rowcolor{gray!8}
R7 & \textbf{Full RIMCA $M_t$} & \textbf{98.00} & 93.80 & \textbf{94.90} & \textbf{94.35} & 1.401 & 50.96 \\
\bottomrule
\end{tabular}
\end{table}

Table~\ref{tab:rimca_layerwise_ablation} shows that the intermediate outputs do not improve step by step. The recurrent state $A_t$ (R4) is the strongest single intermediate readout at 95.17\% mAP$_{50}$ and 92.50 F1. This shows that temporal accumulation is important. The earlier spatial responses and the filtered response $G_t$ are weaker when used alone. R6 improves again when $G_t$ is combined with $A_t$, which supports using both spatial and recurrent temporal information.

The full representation $M_t$ (R7) reaches 98.00\% mAP$_{50}$ and 94.35 F1. Compared with R4, mAP$_{50}$ increases by 2.83 points and recall rises from 90.93 to 94.90, while precision remains similar. This result matches the goal of RIMCA: retain useful temporal responses while reducing background interference. R1--R6 are diagnostic readouts and should not be interpreted as isolated removals of individual operations.

\subsubsection{RIMCA Parameter Sensitivity}
\label{subsec:rimca_sensitivity}


\begin{figure*}[!t]
\centering
\includegraphics[width=0.98\textwidth]{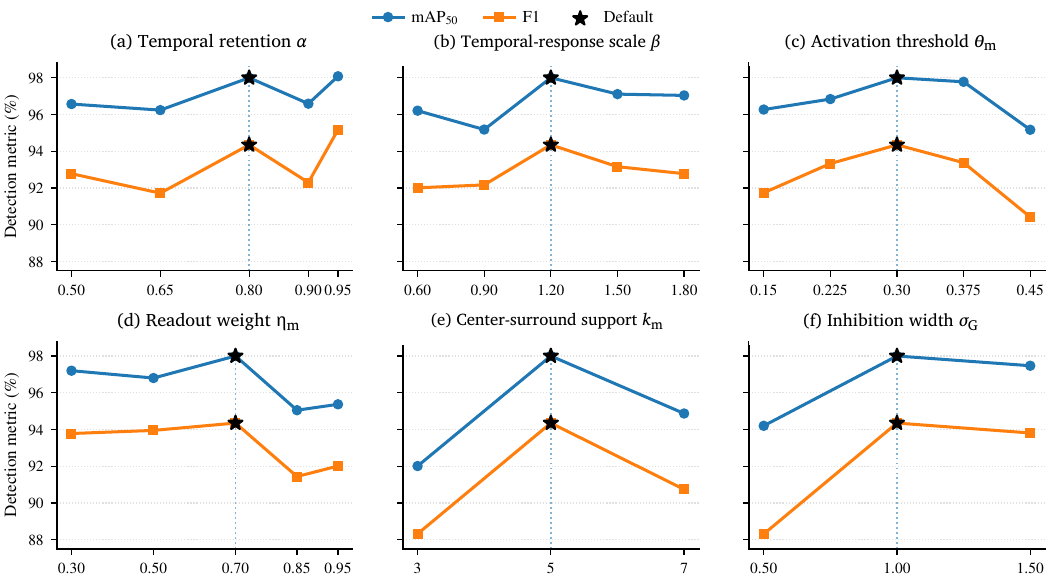}
\caption{One-factor-at-a-time sensitivity analysis of the main RIMCA configuration on DAUB-R. Panels (a)--(f) vary temporal retention $\alpha$, temporal-response scale $\beta$, activation threshold $\theta_m$, readout weight $\eta_m$, center--surround support $k_m$, and inhibition-kernel Gaussian width $\sigma_G$, respectively. The star marks the main-model configuration (98.00\% mAP$_{50}$ and 94.35\% F1); all remaining RIMCA and detector settings are fixed.}
\label{fig:rimca_parameter_sensitivity}
\end{figure*}

Fig.~\ref{fig:rimca_parameter_sensitivity} and Table~\ref{tab:rimca_parameter_sensitivity} vary one RIMCA parameter at a time while keeping the others fixed. The Default column gives the configuration used in all main experiments. We do not select different RIMCA settings for different datasets. The default setting gives strong mAP$_{50}$ and F1 for all six parameter groups in the figure.

Within the tested one-parameter ranges, the strongest visible drops occur when the center--surround support is reduced to $k_m=3$ and when the inhibition width is reduced to $\sigma_G=0.50$. This shows that the spatial scale used for local inhibition is important in the tested range. In comparison, the one-dimensional sweeps of $\alpha$, $\beta$, $\theta_m$, and $\eta_m$ show smaller changes over most of their tested values. The result suggests that RIMCA is less sensitive to moderate changes of these four parameters around the default setting. The main experiments use one shared RIMCA configuration for all three datasets. The default values are not re-selected for each benchmark. Therefore, the reported benchmark results do not rely on a separate RIMCA setting for each dataset. This DAUB-R analysis only describes one-parameter sensitivity around the shared configuration. Because one parameter is changed at a time, it does not measure interactions among parameters.

\begin{table}[!t]
\centering
\caption{RIMCA parameters used in the one-parameter-at-a-time sensitivity analysis. The default column is the main-model configuration.}
\label{tab:rimca_parameter_sensitivity}
\TableFont
\setlength{\tabcolsep}{4.5pt}
\renewcommand{\arraystretch}{1.05}
\begin{tabular}{lcc}
\toprule
\textbf{Parameter} & \textbf{Tested Values} & \textbf{Default} \\
\midrule
$\alpha$ & \{0.50, 0.65, 0.80, 0.90, 0.95\} & 0.80 \\
$\beta$ & \{0.60, 0.90, 1.20, 1.50, 1.80\} & 1.20 \\
$\theta_m$ & \{0.15, 0.225, 0.30, 0.375, 0.45\} & 0.30 \\
$\eta_m$ & \{0.30, 0.50, 0.70, 0.85, 0.95\} & 0.70 \\
$k_m$ & \{3, 5, 7\} & 5 \\
$\sigma_G$ (inhibition width) & \{0.50, 1.00, 1.50\} & 1.00 \\
\bottomrule
\end{tabular}
\end{table}

\subsubsection{Cross-Pathway Feature Interaction}
\label{subsec:interconnection_ablation}

Table~\ref{tab:v14b_convergence_ablation} evaluates Stage~II. It compares appearance-only and motion-only baselines, simple Add/Concat aggregation, LGAG, and PMI. All dual-pathway variants use the same RIMCA representations and RT-DETR decoder.

\begin{table}[!t]
\centering
\caption{Motion--appearance interaction ablation on DAUB-R. All variants use the same RT-DETR decoder, and variants with a motion pathway use the same RIMCA representations. LGAG denotes Large-kernel Grouped Attention Gate~\cite{LGAG2023}. G denotes GFLOPs and Par. denotes millions of parameters.}
\label{tab:v14b_convergence_ablation}
\TableFont
\setlength{\tabcolsep}{1.45pt}
\renewcommand{\arraystretch}{1.04}
\begin{tabular}{@{}lcccccc@{}}
\toprule
\textbf{Variant} & \textbf{mAP$_{50}$} & \textbf{P} & \textbf{R} & \textbf{F1} & \textbf{G} & \textbf{Par.} \\
\midrule
\multicolumn{7}{l}{\textit{Single-pathway baselines}} \\
(1) Appearance-only & 90.20 & 87.50 & 81.70 & 84.50 & 57.20 & 19.88 \\
(2) Motion-only & 90.00 & 90.30 & 83.40 & 86.71 & 57.20 & 19.88 \\
\midrule
\multicolumn{7}{l}{\textit{Non-interactive aggregation}} \\
(3) Element-wise Add & \underline{96.80} & \underline{93.20} & 93.10 & \underline{93.15} & 91.90 & 31.31 \\
(4) Direct Concat & 96.40 & 90.30 & \underline{94.70} & 92.45 & 92.70 & 31.38 \\
\midrule
\multicolumn{7}{l}{\textit{Interactive fusion}} \\
(5) LGAG~\cite{LGAG2023} & 96.50 & 92.80 & 93.50 & \underline{93.15} & 103.20 & 32.20 \\
\rowcolor{gray!8}
(6) \textbf{PMI (Ours)} & \textbf{98.00} & \textbf{93.80} & \textbf{94.90} & \textbf{94.35} & 93.90 & 32.44 \\
\bottomrule
\end{tabular}
\end{table}

The two-pathway variants outperform the single-pathway baselines. Appearance-only and motion-only reach 90.20 and 90.00 mAP$_{50}$, while simple Add and Concat reach 96.80 and 96.40. The motion-only model also gives a higher F1 than the appearance-only model (86.71 versus 84.50). This shows that the RIMCA pathway carries useful target information. The larger gain from Add and Concat further shows that motion and appearance information are complementary even before PMI.

PMI further increases mAP$_{50}$ to 98.00 and F1 to 94.35. Relative to element-wise Add, both metrics improve by 1.20 points. Relative to LGAG, PMI improves mAP$_{50}$ by 1.50 points and F1 by 1.20 points while using fewer GFLOPs. Thus, the motion pathway provides complementary temporal information, and PMI gives an additional gain through bidirectional interaction. This supports the Stage~II design in Sec.~\ref{sec:pmi_encoder}.

\subsubsection{Cross-Backbone Evaluation of the Dual-Pathway Configuration}
\label{subsec:pmi_cross_backbone}

Table~\ref{tab:PMI_generalization} tests the complete RIMCA+PMI dual-pathway design with several YOLO and DETR-style detectors. The +PMI rows add both the motion pathway and PMI, so this table does not isolate PMI alone. The isolated PMI comparison is given in Table~\ref{tab:v14b_convergence_ablation}.

\begin{table}[!t]
\centering
\caption{Cross-backbone evaluation of the dual-pathway RIMCA+PMI configuration on DAUB-R. Var. denotes the evaluated variant; App. and Mot. denote appearance-only and motion-only variants. R18 denotes ResNet-18. G denotes GFLOPs and Par. denotes millions of parameters. The +PMI rows include both the motion pathway and PMI; isolated PMI attribution is given in Table~\ref{tab:v14b_convergence_ablation}.}
\label{tab:PMI_generalization}
\TableFont
\setlength{\tabcolsep}{1.00pt}
\renewcommand{\arraystretch}{1.08}
\begin{tabular}{@{}llcccccc@{}}
\toprule
\textbf{Detector} & \textbf{Var.} & \textbf{mAP$_{50}$} & \textbf{P} & \textbf{R} & \textbf{F1} & \textbf{G} & \textbf{Par.} \\
\midrule

\multirow{3}{*}{YOLOv8~\cite{yolov8}}
& App. & 83.60 & 91.10 & 71.90 & 80.40 & 6.80 & 2.68 \\
& Mot. & 86.90 & 84.70 & 78.90 & 81.70 & 6.80 & 2.68 \\
& +PMI & \textbf{95.80} & \textbf{95.40} & \textbf{93.00} & \textbf{94.20} & 10.50 & 4.32 \\

\midrule
\multirow{3}{*}{YOLOv10~\cite{yolov10}}
& App. & 82.20 & 86.50 & 71.60 & 78.30 & 8.20 & 2.69 \\
& Mot. & 80.90 & 80.90 & 71.80 & 76.10 & 8.20 & 2.69 \\
& +PMI & \textbf{94.10} & \textbf{90.80} & \textbf{90.50} & \textbf{90.60} & 11.70 & 4.01 \\

\midrule
\multirow{3}{*}{YOLOv11~\cite{yolov11}}
& App. & 81.50 & 85.00 & 75.40 & 79.90 & 6.30 & 2.58 \\
& Mot. & 76.40 & 94.90 & 56.40 & 70.80 & 6.30 & 2.58 \\
& +PMI & \textbf{92.90} & 90.30 & \textbf{91.50} & \textbf{90.90} & 10.80 & 3.78 \\

\midrule
\multirow{3}{*}{YOLOv12~\cite{yolov12}}
& App. & 79.70 & 83.90 & 74.50 & 78.90 & 5.90 & 2.53 \\
& Mot. & 87.60 & 86.90 & 78.50 & 82.50 & 5.90 & 2.53 \\
& +PMI & \textbf{97.00} & \textbf{95.80} & \textbf{95.50} & \textbf{95.60} & 10.40 & 5.11 \\

\midrule
\multirow{3}{*}{RT-DETR~\cite{rtdetr}}
& App. & 88.90 & 86.30 & 81.60 & 83.90 & 125.60 & 41.93 \\
& Mot. & 86.00 & 92.20 & 79.70 & 85.50 & 125.60 & 41.93 \\
& +PMI & \textbf{97.20} & \textbf{92.50} & \textbf{92.20} & \textbf{92.30} & 256.40 & 98.62 \\

\midrule
\multirow{3}{*}{MI-DETR (R18)}
& App. & 90.20 & 87.50 & 81.70 & 84.50 & 57.20 & 19.88 \\
& Mot. & 90.00 & 90.30 & 83.40 & 86.71 & 57.20 & 19.88 \\
& +PMI & \textbf{98.00} & \textbf{93.80} & \textbf{94.90} & \textbf{94.35} & 93.90 & 32.44 \\

\bottomrule
\end{tabular}
\end{table}

Across all evaluated detectors, the complete dual-pathway configuration improves mAP$_{50}$ and F1 over the corresponding appearance-only variant. The mAP$_{50}$ increase ranges from 7.8 points for MI-DETR to 17.3 points for YOLOv12. The same trend appears for YOLOv8, YOLOv10, YOLOv11, and RT-DETR, and F1 also improves in every detector family. This shows that the benefit is not tied to one detector design.

The complete configuration is also stronger than the motion-only variant in every group. The relative strength of the two single pathways is not the same for all detectors. For example, motion-only is stronger for YOLOv12, while appearance-only is stronger for YOLOv11. However, combining the two pathways gives the best mAP$_{50}$ and F1 in both cases. This supports the complementary role of appearance and motion information. The trend is consistent for both YOLO-style and DETR-style detectors. The extra pathway and interaction add parameters and computation, as shown in the last two columns. Because the +PMI rows change both the pathway and the interaction module, this table does not isolate PMI. Table~\ref{tab:v14b_convergence_ablation} provides that controlled comparison.


\section{Conclusion}
We presented MI-DETR for moving infrared small target detection. RIMCA reduces local background responses and recurrently integrates frame-to-frame temporal responses into a causal, pixel-aligned motion representation, while PMI performs bidirectional interaction between the separate motion and appearance pathways before RT-DETR detection. Experiments on DAUB-R, ITSDT-15K, and IRDST-H show the highest mAP$_{50}$ among the compared methods on all three benchmarks. Controlled motion-representation, RIMCA, PMI, sensitivity, and cross-backbone analyses further support the roles of the proposed components. The complete pipeline reaches 34.60 FPS under the unified RTX~3090 protocol while achieving 70.30\% mAP$_{50}$ on IRDST-H.

\bibliographystyle{IEEEtran}
\bibliography{egbib}

\raggedbottom
\makeatletter
\def\@IEEEBIOskipN{1.5\baselineskip}
\makeatother

\end{document}